# Determinants of Training Corpus Size for Clinical Text Classification

Jaya Chaturvedi[1*†], Saniya Deshpande[1], Chenkai Ma[1], Robert Cobb[1], Angus Roberts[1], Robert Stewart[1], Daniel Stahl[1], Diana Shamsutdinova[1†]

[1]King's College London, United Kingdom

[†]**Joint first authors**

**\* Correspondence:**
Corresponding Author
jaya.chaturvedi@kcl.ac.uk

Abstract

**Introduction:** Clinical text classification using natural language processing (NLP) models requires adequate training data to achieve optimal performance. Despite large language models (LLMs) offering zero-shot solutions for extracting information from clinical texts, fine-tuning remains necessary in many specialist cases or when superior accuracy is prioritized. For that, 200-500 documents are typically annotated. The number is constrained by time and costs and lacks justification of the sample size requirements and their relationship to text vocabulary properties.

**Methods:** Using the publicly available MIMIC-III dataset containing hospital discharge notes with ICD-9 diagnoses as labels, we employed pre-trained BERT embeddings followed by Random Forest classifiers to identify 10 randomly selected diagnoses, varying training corpus sizes from 100 to 10,000 documents, and analyzed vocabulary properties by identifying strong and noisy predictive words through Lasso logistic regression on bag-of-words embeddings.

**Results:** Learning curves varied significantly across the 10 classification tasks despite identical preprocessing and algorithms, with 600 documents sufficient to achieve 95% of the performance attainable with 10,000 documents for all tasks. Vocabulary analysis revealed that more strong predictors and fewer noisy predictors were associated with steeper learning curves, where every 100 additional noisy words decreased accuracy by approximately 0.02 while 100 additional strong predictors increased maximum accuracy by approximately 0.04.

**Discussion:** Pre-trained BERT embeddings enable effective clinical text classification with as few as 600 gold standard annotations. Understanding the relationship between vocabulary quality and learning curve shapes can guide optimal sample size decisions, suggesting that keyword analysis can predict model performance and that data cleaning techniques focusing on document relevance may be more valuable than simply increasing training corpus size, though limitations include potential bias from random ICD-10 condition selection and co-morbidity effects.

# 1    Introduction

Appropriate size of a training data sample is essential for quantitative research, especially for the generalizability and reproducibility of the findings (Okul et al., n.d.; Ehrenberg 2000). Knowing whether an appropriate sample size was used is crucial to determine the value of a research project. This is because it gives an insight into whether appropriate considerations have been made to ensure the project is ethical and methodologically sound (Faber and Fonseca 2014). It is also important to use appropriate methods for creating such samples, such as random sampling, in order to avoid magnifying unrepresentativeness within large samples (Msaouel et al. 2023). Yet, a study found that 60% of publications do not perform sample size calculations or provide sufficient details on sampling methods (Okul et al., n.d.).

Research methods that utilize natural language processing (NLP) are no exception. NLP is a branch of artificial intelligence (AI) within computer science that combines computational linguistics with statistical and machine learning methods, enabling the analysis and processing of text data (Liddy 2001). NLP methods are widely used within healthcare research due to the growing volume of data available from EHR databases (Esteva et al., 2019). EHRs within a single hospital could generate about 150,000 pieces of data (Esteva et al., 2019), many of which may be in textual form. Text is often used for ease of capturing fine-grained details or supplemental information which do not fit into any predetermined structured fields, such as patients' medical histories, preliminary diagnoses, and medications (Ehrenstein et al., 2019)

A prerequisite for a good sample size is the availability of sufficient data to represent the larger population and provide adequate precision in output. This can be quite challenging in the healthcare domain due to the scarcity of openly available healthcare datasets, and the privacy regulations surrounding the use of such data from hospitals and primary care. Data scarcity is compounded by the fact that NLP-based supervised machine learning models require large numbers of human-annotated (in the healthcare domain, ideally clinician-annotated) data as a prerequisite, which can be a limitation due to time and cost constraints (Negida et al., 2019). Insufficient sample sizes within NLP can lead to algorithms that do not perform adequately. This highlights the importance and need for guidelines on what sample sizes are needed, given limited data, to build machine learning models that perform well.

When referring to machine learning models in general and within NLP, sample size calculations can be used at different stages, such as for training, validation, and testing. Previous work has been conducted to determine sample sizes for validation (Negida et al., 2019; Riley et al., 2020) and limited research has been published on general sample sizes for NLP (Sordo et al., 2005). Sordo et al. (2005) examine the effect of sample size on the accuracy of classification with three classification methods (Naive Bayes, Decision Trees, and Support Vector Machines) using narrative reports from a hospital, classifying the smoking status of patients (Sordo et al., 2005). They conclude that there is indeed a correlation between the size of the training set and the classification rate, and models show improved performance when they are trained with bigger samples (Sordo et al., 2005). Using large electronic health record (EHR) databases, a recent study by Liu et al. (2021) highlighted the importance of selecting a sample that is unbiased and truly representative of the population in order to ensure high quality research (Liu et al., 2021). Furthermore, despite a very large corpus of texts being used for large language models training, such as Chat-GPT, Llama, or BERT, there is little insight into the minimum size of the post-training datasets, if post-tuning is used. While our work does not address issues of bias due to the methodologies used to select samples, it does aim to extend previous research and explore the impact of sample sizes as well as vocabulary quality for a binary classification task.

Classification tasks using natural language processing (NLP) models are often used in clinical research to automate the identification of specific information in EHRs. Such information can be a specific diagnosis (e.g., diabetes, or depression), a symptom (e.g. feeling of pain), or a treatment (e.g. therapy, prescription). This study aims to explore sample size requirements for training data for classification models using clinical text, and how such requirements depend on the language properties of the underlying documents. The practical question we address in this project is whether the 200-500 documents that usually get annotated in clinical research studies are enough.

In our experiments, we took advantage of pre-trained Large Language Models (LLMs) for numerical representation of clinical text and then trained tree-based and neural net (NN) based classifiers. Pre-trained language models like BERT (Devlin et al., 2019) were chosen for their ability to capture contextual relationships in clinical text through transfer learning (an approach where the knowledge learnt in one task or domain is applied/transferred to solve another task). These representations were paired with Random Forest classifiers which work well with small sample sizes and provide better interpretability. We aimed to investigate how much training data an LLM-augmented text classification model needs to learn a generalizable function.

## 2 Materials and Methods

### 2.1 Data Source

Medical Information Mart for Intensive Care (MIMIC-III) is an EHR database which was developed by the Massachusetts Institute of Technology (MIT) and made available for researchers under a specified governance model (Johnson et al., 2016). MIMIC-III contains data on over 50,000 hospital admissions for over 38,000 patients, including about 1.2 million de-identified clinical notes, such as nursing and physician notes, discharge summaries, and ECG/radiology reports (Johnson et al., 2016).

MIMIC-III was chosen for this study due to ease of access, thereby making the study reproducible. MIMIC-III is commonly used in healthcare research (Mayaud et al., 2013; Lehman et al., 2012; Velupillai et al., 2015; Abhyankar et al., 2014).

### 2.2 Ethics and Data Access

Access to the MIMIC-III database requires that the data be handled with care and respect as it contains detailed information about the clinical care of patients. Access was formally requested and granted through the processes documented on the MIMIC-III website. A course protecting human research participants, including the requirements outlined by HIPAA (Health Insurance Portability and Accountability Act) was completed. A data use agreement outlining appropriate data usage and standards of security was submitted.

### 2.3 Data Selection

From MIMIC-III (Johnson et al., 2016), we selected 10 randomly chosen diagnoses codes, to represent diverse classification tasks. For each diagnosis, we extracted the hospital discharge notes and their corresponding ICD-9 diagnoses. These are listed in table 1. Since the process of manual annotation would be time consuming, we used the ICD-9 diagnosis codes as labels for the documents. Google BigQuery was used to extract this data using SQL queries.

| ICD-9 Code | Diagnosis | Count |
|---|---|---|
| 401.9 | unspecified essential hypertension | 20,703 |
| 530.81 | esophageal reflux | 6,326 |
| 518.81 | acute respiratory failure | 7,497 |
| 427.31 | atrial fibrillation | 12,891 |
| 414.01 | coronary atherosclerosis of native coronary artery | 12,429 |
| 250.00 | diabetes mellitus without mention of complication, type II or unspecified type, not stated as uncontrolled | 9,058 |
| 599.0 | urinary tract infection, site not specified | 6,555 |
| 584.9 | acute kidney failure, unspecified | 9,119 |
| 428.0 | congestive heart failure, unspecified | 13,111 |
| 272.4 | hyperlipidemia, unspecified | 8,690 |

**Table 1. List of 10 randomly chosen diagnosis codes**

## 2.4 Classification

For each ICD-9 diagnosis code (for example, 401.9 - label 1), a random stratified subset was selected (not 401.9 - label 0). As we only focused on classification, we took an encoder-only BERT model (Devlin et al., 2019) and extracted pooled level embedding vectors for all the documents in the training corpus. Pre-trained BERT models were chosen for their ability to capture contextual relationships inherent in clinical narratives. Unlike traditional approaches such as TF-IDF or bag-of-words, BERT's contextual embeddings can distinguish nuanced meanings where identical terms have different clinical significance based on context. Specifically for our research question, transfer learning from pre-trained models allowed us to leverage knowledge from a vast number of documents during the pre-training of the model, potentially compensating for when the samples had limited annotated training data. This was particularly relevant when investigating whether 200-500 docs provide sufficient information for training, as the model begins with a rich representation rather than starting from scratch.

Our pilot investigation explored approaches such as zero-shot learning (Chat-GPT 3.5 and LLama2 7B), question-answering formats using DistilBERT and BERT_Large, as well as GPT-2 with various prompting strategies. However, these approaches showed low performance, also reported by others (Yuting et al., 2024).

Training BERT's final layer for classification directly showed an inferior performance compared to a baseline Random Forest classifier. Hence, we focused on the pipeline where pre-trained BERT embeddings were followed by the Random Forest Classifier. We trained classification models to identify one of the 10 randomly chosen diagnoses. The size of the training corpus, N, varied from 100 to 10,000 to investigate how fast the performance improves with an increase in the training corpus size. For each N, training datasets were randomly drawn 10 times, the resulting models were validated in a large hold-out dataset (5000) and the averaged performance metrics reported.

We further hypothesized that the size of the training corpus might be associated with the vocabulary properties of the underlying clinical texts, in particular, with the number of strong and noisy words in the documents. Strong and noisy words were identified by running a Lasso logistic model on a simple bag-of-words embedding to predict document class and defining noisy and strong words based on their regression coefficients. We employed SHAP (SHapley Additive exPlanations) linear explainer to quantify feature contributions for additional interpretability insights. Features with absolute coefficient values >0.1 were classified as "strong predictors," while those below this threshold were considered potential noise. We tested the association of the vocabulary properties with classification performance at 10000 and with the learning curve steepness (performance difference at 10000 and 300 training documents) for the 10 diagnoses.

The code has been made available on GitHub[1].

## 3    Results

Vocabulary sizes for each selected ICD-9 diagnosis and their stratified random counterpart are shown in figure 1.

---

[1] https://github.com/dianashams/NLP_sample_size_simulation_study/

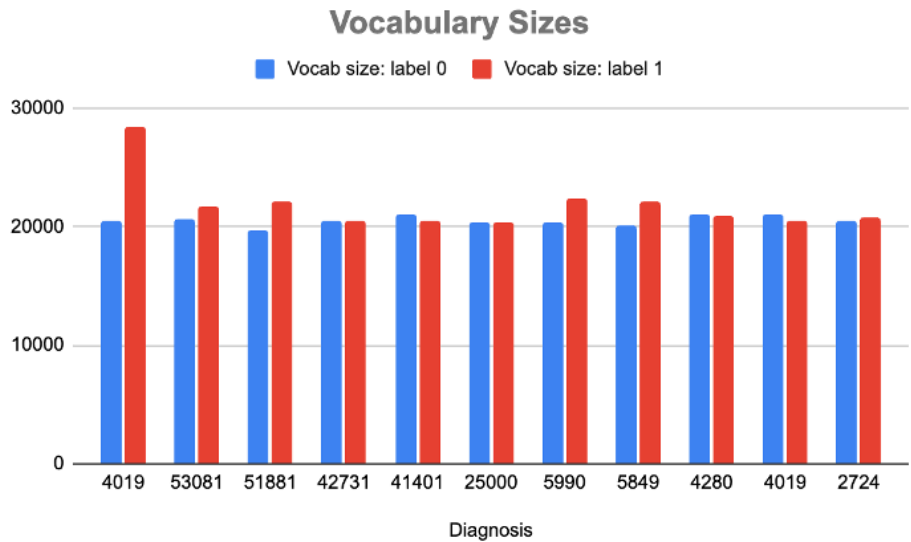

**Figure 1. Vocabulary sizes for the 10 diagnosis codes**

Learning curves of the 10 classification tasks differed significantly despite using the same preprocessing pipeline, classification algorithms, and documents' source. In terms of practical guidance, although 100-300 documents were often suboptimal, 600 were enough to achieve 95% of the performance that would have been reached with 10,000 documents for all 10 tasks (Figure 2).

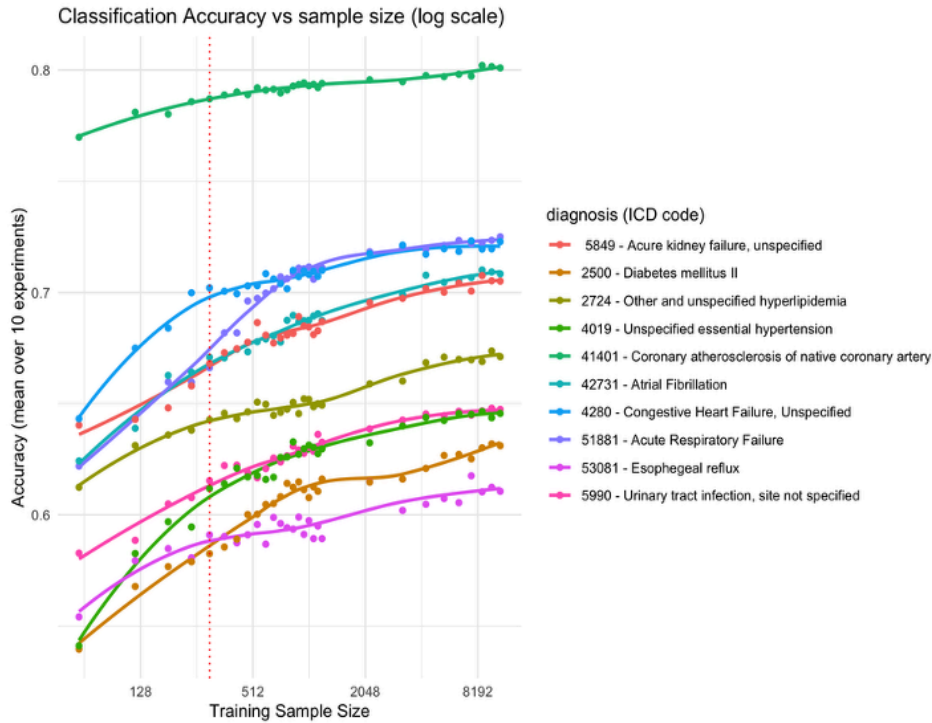

**Figure 2. Learning curves for binary classification (diagnostic code identification), accuracy. [dotted red line indicates performance at N=600]**

Analyses of the text vocabularies (Figure 3) showed a clear pattern between predictor quality and potential classification accuracy: more strong predictors and fewer noisy predictors were associated with steeper learning curves, and vice versa. Based on the analysis excluding the hypertension dataset (an outlier), an increase of 100 noisy words corresponded to a decrease of approximately 0.02 in the accuracy at the largest training corpus size, 10000. In addition, an increase of 100 strong predictors corresponds to an increase of approximately 0.04 in the maximum accuracy.

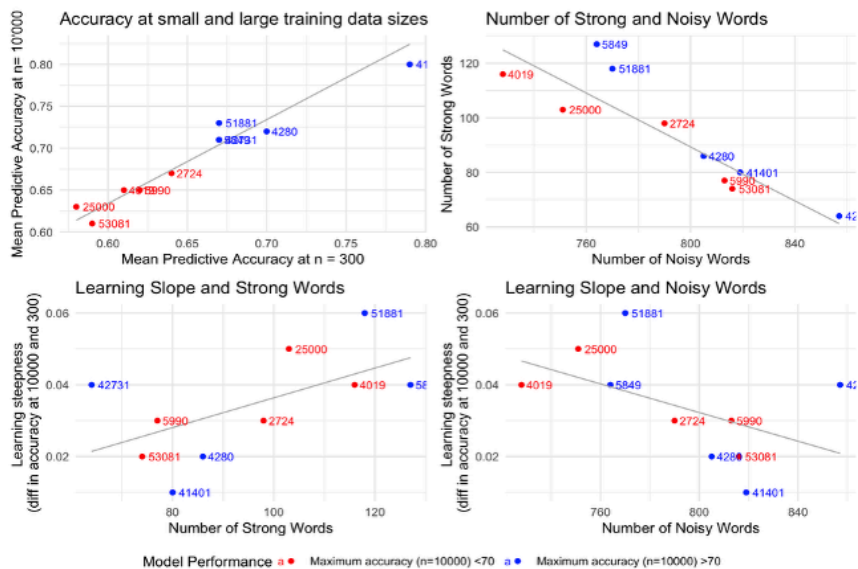

**Figure 3. Learning curve analysis: the relationship between the number of strong predictive words, noise words, and the learning curve steepness.**

The SHAP analysis for the diabetes classification example (Figure 4) demonstrates that models achieve robust clinical understanding with just 1,000 training samples. Both the 1,000 and 12,000 sample models successfully identify the core diagnostic terms ('diabetes', 'metformin', 'dm'), with these terms appearing at the top of both SHAP importance rankings. While the larger training set provides marginal refinements in feature ordering and slightly cleaner separation of clinical terms from generic words, the 1,000-sample model already captures the essential clinical concepts needed for accurate classification. The presence of some generic terms ('with', 'and', 'hospital') in the smaller model's feature set does not compromise its fundamental understanding of diabetes-related vocabulary. This validates our vocabulary analysis findings that tasks with many noisy predictors require more training data not to learn new concepts, but to filter out non-discriminative features. The 600-1,000 document range could represent where models learn essential clinical vocabulary, and that additional data primarily serves to reduce noise rather than capture novel clinical insights.

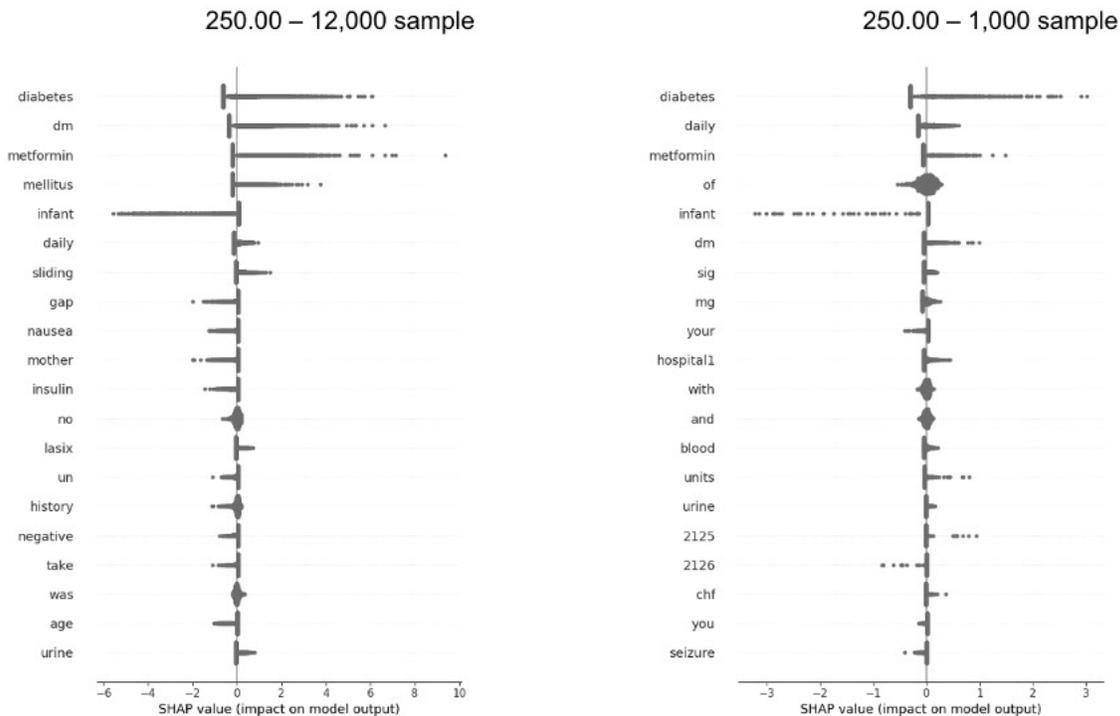

**Figure 4. SHAP for diagnosis code 250.00**

## 4 Discussion

We found that using contextual LLM embeddings combined with Random Forest for clinical text classification can be effective, and that one may not need more than 600 gold standard annotations for classification tasks. Further, our work shows that understanding the relationship between strong predictors, noisy data, and accuracy can guide decisions on optimal sample sizes for training data in NLP tasks, potentially saving computational resources and time. In particular, a simple analysis of keywords can serve as an indicator of the learning curve's steepness. Instead, data cleaning techniques focused on identifying and isolating relevant parts of larger documents can significantly impact model performance. This insight can be valuable for estimating the potential performance of NLP models before extensive training.

Our findings align with recent work demonstrating the limitations of zero-shot and few-shot learning for clinical tasks (Yuting et al., 2024). While LLMs show impressive capabilities in the general domain, tasks such as clinical classification require more domain-specific fine-tuning. Recent studies (Bucher at al., 2024) have found that fine-tuning outperforms zero- and few-shot approaches in clinical research.

Our results highlight a number of practical implications. When possible, researchers should plan for at least 600 annotations, especially when dealing with complex diagnostic categories that might have noisy vocabularies. A preliminary vocabulary analysis on small, annotated samples could inform the need for larger annotation efforts. In addition, data cleaning and document segmentation to isolate relevant content might be more impactful than increasing the training sample size. The 600-document threshold represents the sample size needed for typical clinical heterogeneity. Tasks with unusually homogeneous (high cosine) or heterogeneous (low cosine) documentation may require adjusted sample sizes, which can be predicted through preliminary similarity analysis.

A limitation of this work is the random selection of ICD-10 conditions, which may have introduced some bias since some conditions frequently occur as co-morbidities. A more systematic selection of conditions could strengthen this analysis. Further work can include comparisons of different types of human-machine annotations (Cardamone et al., 2025; Mustafa et al., 2025) and investigations of their learning speed and dependence on text properties and data quality, less covered in current research. In addition, future work will investigate the effect of sample document lengths beyond the current analysis of noisy words. Validation on additional datasets, and extension to multi-class classification would also strengthen the generalizability of this work.

Rather than using LLMs directly for classification, future work should explore LLMs as annotation assistants, generating initial labels for human review. This could help reduce annotation time while maintaining quality.

We have provided empirical evidence that 600 gold-standard annotations represent a practical threshold for achieving near-optimal performance in clinical text classification tasks. The strong relationship between vocabulary properties and learning efficiency offers an effective framework for annotation planning. As the field moves toward human-AI collaborations, these insights can guide the development of more efficient and effective clinical NLP.

## 5 Acknowledgments

The research team declares no conflicts of interest and uses data that is openly available upon completion of privacy certifications with the provider. The project is funded by the ECR Seed Funding Scheme, King's College London, London, and by the NIHR Maudsley Biomedical Research Centre at South London and Maudsley NHS Foundation Trust and King's College London, UK. All code is available on GitHub. The views expressed are those of the author(s) and not necessarily those of the NIHR or the Department of Health and Social Care.

## 6 Conflict of Interest

RS declares research support received in the last 3 years from Janssen, GSK and Takeda. The other authors declare that the research was conducted in the absence of any commercial or financial relationships that could be construed as a potential conflict of interest.

## 7 Funding

AR was funded by Health Data Research UK, an initiative funded by UK Research and Innovation, Department of Health and Social Care (England) and the devolved administrations, and leading medical research charities. DS and RS are part-funded by the National Institute for Health Research (NIHR) Biomedical Research Centre at South London and Maudsley NHS Foundation Trust and King's College London. RS is additionally part-funded by the National Institute for Health Research (NIHR) Applied Research Collaboration South London (NIHR ARC South London) at King's College Hospital NHS Foundation Trust, and by the DATAMIND HDR UK Mental Health Data Hub (MRC grant MR/W014386). JC was supported by the Maudsley Charity. DSh is funded by the King's College London Biostatistics and Health Informatics PhD studentship. The funders had no role in study design, data collection and analysis, decision to publish, or preparation of the manuscript.

## 8 Author Contributions

DS, JC, RC, SD and CM conducted the analysis. DS and JC drafted the manuscript. All authors contributed to reviewing and editing the manuscript.

## 9 Data Availability Statement

The datasets analyzed for this study are openly available through PhysioNet. All code is available on GitHub at https://github.com/dianashams/NLP_sample_size_simulation_study/

Liu L, Bustamante R, Earles A, Demb J, Messer K, Gupta S. A strategy for validation of variables derived from large-scale electronic health record data. Journal of Biomedical Informatics [Internet]. 2021 Sep 1 [cited 2022 Aug 4];121:103879. Available from: https://www.sciencedirect.com/science/article/pii/S1532046421002082

Johnson, A., Pollard, T., Shen, L. et al. MIMIC-III, a freely accessible critical care database. Sci Data 3, 160035 (2016). https://doi.org/10.1038/sdata.2016.35

Mayaud L, Lai PS, Clifford GD, Tarassenko L, Celi LAG, Annane D. Dynamic data during hypotensive episode improves mortality predictions among patients with sepsis and hypotension. Crit Care Med [Internet]. 2013 Apr [cited 2022 Aug 4];41(4):954–62. Available from: https://www.ncbi.nlm.nih.gov/pmc/articles/PMC3609896/

Lehman L wei, Saeed M, Long W, Lee J, Mark R. Risk stratification of ICU patients using topic models inferred from unstructured progress notes. AMIA . Annual Symposium proceedings / AMIA Symposium AMIA Symposium [Internet]. 2012;2012:505–11. Available from: /pmc/articles/PMC3540429/?report=abstract

Velupillai S, Mowery D, South BR, Kvist M, Dalianis H. Recent Advances in Clinical Natural Language Processing in Support of Semantic Analysis. Yearb Med Inform [Internet]. 2015 [cited 2022 Aug 4];24(1):183–93. Available from: http://www.thieme-connect.de/DOI/DOI?10.15265/IY-2015-009

Abhyankar S, Demner-Fushman D, Callaghan FM, McDonald CJ. Combining structured and unstructured data to identify a cohort of ICU patients who received dialysis. Journal of the American Medical Informatics Association [Internet]. 2014 Sep 1 [cited 2022 Aug 4];21(5):801–7. Available from: https://doi.org/10.1136/amiajnl-2013-001915

Devlin J, Chang MW, Lee K, Toutanova K. BERT: Pre-training of Deep Bidirectional Transformers for Language Understanding. In: Proceedings of the 2019 Conference of the North American Chapter of the Association for Computational Linguistics: Human Language Technologies, Volume 1 (Long and Short Papers) [Internet]. Minneapolis, Minnesota: Association for Computational Linguistics; 2019 [cited 2021 Apr 21]. p. 4171–86. Available from: https://www.aclweb.org/anthology/N19-1423

Yuting Guo, Anthony Ovadje, Mohammed Ali Al-Garadi, Abeed Sarker, Evaluating large language models for health-related text classification tasks with public social media data, Journal of the American Medical Informatics Association, Volume 31, Issue 10, October 2024, Pages 2181–2189, https://doi.org/10.1093/jamia/ocae210

Cardamone, Nicholas C., et al. "Classifying Unstructured Text in Electronic Health Records for Mental Health Prediction Models: Large Language Model Evaluation Study." JMIR Medical Informatics 13.1 (2025): e65454.

Mustafa, A., Naseem, U., & Azghadi, M. R. (2025). Large language models vs human for classifying clinical documents. International Journal of Medical Informatics, 105800.11